\documentclass[letterpaper,12pt]{article}

\usepackage{mathptmx}
\usepackage{amsmath}

\usepackage[colorlinks=true,allcolors=blue]{hyperref}

\usepackage{graphicx}
\graphicspath{{figures/}}

\usepackage{multirow}
\usepackage[comma,sort&compress,square,numbers]{natbib}

\usepackage{lineno}[displaymath, mathlines]

\usepackage{censor}

\usepackage{xcolor}

\begin{document}



\begin{center}
    \Large{Method and Validation for Optimal Lineup Creation for Daily Fantasy Football Using Machine Learning and Linear Programming}
\end{center}

\vspace{0.15in}

    \begin{center}
          Joseph M.~Mahoney
          \footnote{Mechanical Engineering, Alvernia University, Reading, PA, USA} 
          \footnote{Mechanical Engineering, The Pennsylvania State University, Berks College, Reading, PA, USA}
          \footnote{Kinesiology, The Pennsylvania State University, Berks College, Reading, PA, USA}
          \footnote{\href{mailto:joseph.m.mahoney@gmail.com}{joseph.m.mahoney@gmail.com}}
        \hspace{0.75in}  Tomasz B.~Paniak\footnotemark[2]{}
\end{center}

\section*{Abstract}

Daily fantasy sports~(DFS) are weekly or daily online contests where real-game performances of individual players are converted to fantasy points~(FPTS). Users select players for their lineup to maximize their FPTS within a set player salary cap. This paper focuses on~(1) the development of a method to forecast NFL player performance under uncertainty and~(2) determining an optimal lineup to maximize FPTS under a set salary limit. A supervised learning neural network was created and used to project FPTS based on past player performance~(2018 NFL regular season for this work) prior to the upcoming week. These projected FPTS were used in a mixed integer linear program to find the optimal lineup. The performance of resultant lineups was compared to randomly-created lineups. On average, the optimal lineups outperformed the random lineups. The generated lineups were then compared to real-world lineups from users on DraftKings. The generated lineups generally fell in approximately the 31\textsuperscript{st} percentile~(median). The FPTS methods and predictions presented here can be further improved using this study as a baseline comparison.

\newpage
\section{Introduction}\label{sec:intro}

Since 2003, fantasy sports have seen a 290\% increase in users~\cite{Industry10:online}. There is a growing market for users to take on the role of a team manager by creating virtual sports teams and competing online. Users may take part in contests that span an entire season, a week, or even a single game. Daily fantasy sports~(DFS) are typically weekly or daily contests that allow users to create a lineup of players before the start of a real game. Real-game statistics of players are converted to a scoring system known as fantasy points~(FPTS). The summation of selected players' FPTS are ranked among competing lineups, and the top 1\% to 50\% of users may receive a payout. Multiple popular online platforms cater to DFS, including DraftKings\textsuperscript{\textregistered}, Fan~Duel\textsuperscript{\textregistered}, and
Yahoo!~Sports\textsuperscript{\textregistered}. 
Each platform features several contest styles as well as moderately different scoring systems and rules. Users may enter contests for leagues such as the National Football League~(NFL), Major League Baseball~(MLB), and the National Basketball Association~(NBA). For this paper, ``player'' is used to refer to the athletes competing in the real-world game while ``user'' is employed to refer to the people generating the fantasy drafts and entering them into ``contests.'' 

DFS users strive to create the best possible lineup of players before the start of a contest. However, unlike other forms of fantasy sports that commonly feature a preseason draft that allows a player to be on only one user's lineup, DFS allows a player to be on multiple users' lineups. For \textit{most} DFS, each player is assigned a salary and users have a salary cap restriction on their lineup. As with traditional fantasy sports, there are a specific number of positions that must be filled on the lineup. Generating a DFS lineup can thus be posed as a resource allocation optimization problem: specifically, a Stochastic Knapsack Problem~(SKP). A \textit{general} Knapsack Problem is a problem of filling a knapsack with the most valuable combination of objects~--~each with its own value and volume~--~without exceeding the allowable volume of the knapsack. An SKP adds further complexity as here, the value of each object is uncertain before it is selected for inclusion in the knapsack. In DFS, the lineup can be thought of as a knapsack with its ``volume'' determined by both the salary cap and required player positions. The available players are the ``objects'' with their value determined by their FPTS and their ``volume'' as their salary and position. Since the lineup is set before the start of the game, players' FPTS are uncertain prior to the knapsack being packed~\cite{newell2017optimizing}.


Predicting the number of FPTS that a player will earn~(among other human performance metrics) is a difficult, multi-factorial endeavor. Machine learning techniques, specifically neural networks~(NN), have been applied previously in a range of sports predictions. Neural networks have been previously used for simulations as a comparison against recruiting managers in recruiting new rugby players~\cite{mccullagh2010data}, predicting the outcome of NBA~\cite{loeffelholz2009predicting} and NFL~\cite{david2011nfl} games, predicting the performance of young~\cite{silva2007use} and Olympic swimmers~\cite{edelmann2002modeling}, recruiting javelin throwers~\cite{maszczyk2011neural}, predicting javelin flight~\cite{maier2000neural}, and predicting the performance of cricket players in upcoming games~\cite{iyer2009prediction}. The predictive power of NN is limited by the data available and the relationships among the input and output data. However, in the past, data analysis using NN has produced results capable of predicting trends in athletes and teams. 

Specifically in football, previous work has been done on lineup generation for season-long and daily fantasy contests. One DFS study developed multiple statistical models to predict NFL quarterback FPTS using a backwards stepwise regression, support-vector machine regression, regression tree, random forests, boosting, artificial neural networks, and principal components regression finding that some models predicted FPTS more accurately than websites people rely on for lineup selection~\cite{king2017predicting}. This work shows a promise for determining player value using machine learning paired with other statistical modeling methods. However, the approach determined only quarterback FPTS without addressing the FPTS of the remaining lineup positions. This limits its application in DFS contests in selecting a weekly lineup. Unlike DFS drafts, a season-long NFL fantasy contest requires users to select and draft a lineup that is used week-over-week during the season. One study developed an analysis of historical data from multiple seasons to determine player value and using a mixed integer program found an optimal lineup~\cite{becker2016analytical}. Another study of the seasonal NFL fantasy draft considered a heuristic approach by assuming the opponent's position needs and drafting strategies to select the most valuable lineup~\cite{fry2007player}. A DFS draft, however, requires weekly FPTS predictions that adapt as the season progresses.



Selecting players for an optimal lineup has been previously done with integer linear programming~\cite{newell2017optimizing, eastondaily, newell2017optimizing2, belien2013optimization}. In these works, models assumed a normal distribution of a player's FPTS and that player FPTS were independent of one another. These assumptions simplified the application of integer program to optimize lineups. However, for example, the number of receptions a wide receiver has is directly linked to the number of completions a quarterback throws: there is certainly FPTS correlation among players on the same team that cannot be neglected. One integer program~\cite{newell2017optimizing2, newell2017optimizing} focused on maximizing the expected payout and maximizing the expected FPTS while applying constraints on the lineup.
Similar studies conducted by the same authors submitted lineups to MLB DFS contests and simulated NFL DFS contests by generating random and optimized lineups using an integer program maximizing expected FPTS~\cite{eastondaily, newell2017optimizing}. Another optimization study used post-game cycling stats to select an optimal lineup using mixed integer programming as a post comparison against the winning fantasy lineup~\cite{belien2013optimization}. However, this integer program would not be applicable to a pre-game scenario of selecting the most valuable athletes \textit{before} a contest. 


These previous works have not shown a method and real-world testing for generating optimal weekly lineups for DFS. This research focuses on developing a methodology to~(1) determine the expected performance of players based on objective historical data~--~here, from the 2018 regular NFL season~--~and~(2) generate an optimal lineup that maximizes expected FTPS for the upcoming week's DFS contest. This paper considers NFL Guaranteed Prize Pool~(GPP) ``classic'' style contests where lineups are selected from a list of eligible players for the upcoming Thursday through Monday games. GPP refers to contests with a set entry fee, and users get a share of a predetermined prize pool that may pay out to the top 20\% of lineups. Each player is assigned a position~(i.e.,~Quarterback~(QB), Running Back~(RB), Wide Receiver~(WR), Tight End~(TE) and Defensive Special Team~(DST)) as well as a salary by DraftKings. Note that DST is a \emph{team's} entire defense and special teams. Also, note the salary is determined weekly by DraftKings and is not equal on a player's real-world salary. Instead, a player's salary is influenced, in part, by their previously-earned FPTS and their selection popularity among users in previous weeks~\cite{DraftKin56:online}. DraftKings' GPP ``classic'' style format enforces a \$50,000 salary cap and a lineup constructed of 1~QB, 2~RB, 3~WR, 1~TE, 1~DST and a ``Flex'' player. The Flex player can be a RB, WR or TE. However, this is not meant to imply that this is the \textit{only} style of contest or platform on which the methodology could be applied. Players earn FPTS based on their position and real-world performance in the week's game. For example, a WR will receive 1 point per reception, 0.1 points per yard before or after the catch and 6 points for a touchdown. The full formula for earning points is enumerated on the \href{https://www.draftkings.com/help/rules/1/1}{DraftKings website}, however, the specifics are not pertinent for this method.

Objective player and team data were collected~(Sec.~\ref{sec:agg}) and used to train a NN to predict weekly player FPTS~(Sec.~\ref{sec:exp}). The predicted FPTS of players was used in a linear program to generate the optimal lineup expected to maximize total FPTS~(Sec.~\ref{sec:milp}). The generated lineups were then compared to randomly-created lineups~(Sec.~\ref{sec:random}) and real-world user lineups~(Sec.~\ref{sec:real}). The performance in the real-world contests can be used as a baseline comparison for future improvements to this methodology~(Sec.~\ref{sec:conc}) or competing methods. 

\section{Methods}\label{sec:methods}

\begin{figure}[t!!]
 \centering
 \includegraphics[width=.66\linewidth]{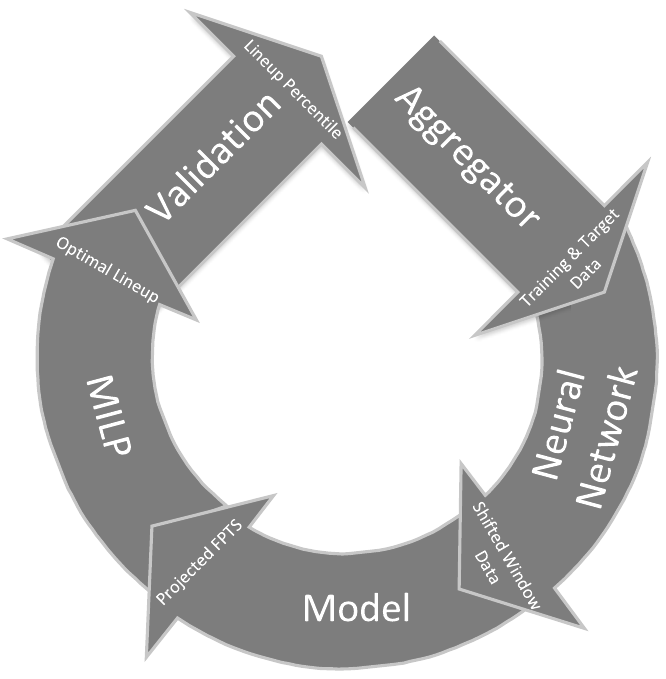}
 \caption[Process Flowchart]{The forecasting of player FPTS and generating a lineup is a four-step process followed by lineup performance validation. \textit{Aggregator}: The compiling and sorting of relevant historical player and team data for training the NN. \textit{Neural Network}: The training of a network to predict NFL player FPTS using data aggregated. \textit{Model}: Once the NN stops improving its output, it creates a model with which new features may be used to predict player FPTS. \textit{MILP}: Player predicted FPTS are used with constraints defined by DraftKings to find a lineup that maximizes FPTS while staying within the constraints. \textit{Validation}: The optimized lineup performance is compared to randomly generated lineups as well as real-world contest lineups.}
 \label{fig:flow}
\end{figure}

The generation of a lineup follows a four-step process of~(1) collecting and aggregating player data~(Sec.~\ref{sec:agg}), (2) training a Neural Network~(NN) to project FPTS~(Sec.~\ref{sec:exp}), (3) using the trained NN to project player FPTS in the upcoming games, and (4) using mixed integer linear programming~(MILP) to generate an optimal lineup that stays within contest's salary and position constraints~(Sec.~\ref{sec:milp}). 
The entire process is shown as a flowchart in Fig.~\ref{fig:flow}. All computations, simulations and statistical analyses used in this paper were performed in Matlab~(R2018b, Natick, MA). 

\subsection{Data collection \& aggregation}\label{sec:agg}

Features believed to be relevant predictors of player's upcoming performance were selected \textit{a priori} and used consistently over the season. Each position has unique ways to earn FPTS, and some positions are likely to earn more FPTS than others such as a quarterback that is involved in nearly all offensive plays. A player's past weeks' FPTS may be a first good indication of future performances~\cite{Whichsta92:online}. Point differentials (the difference in points scored by the player's team to points scored by the opposing team) may be positively correlated to a player's earned FPTS for that game. Similarly, observing offensive- and defensive rankings may predict how well a player's team is expected to do against the opposing team's offense or defense~\cite{boulier2003predicting}. 

The location of the game is also expected to affect the player's psychological state and performance~\cite{bray2000athletes, terry1998influence}. During a home game, a player has home advantage including more fans and a familiar environment. During an away game, the player is expected to be at a disadvantage. The latitudes and longitudes of games or direction of travel may affect a player's circadian rhythm~\cite{worthen1999direction}. Likewise, playing conditions such as a change in elevation or climate may affect a player's performance. 

The spread and over/under are both representations of how well teams are expected to do according to professional oddsmakers. Betting lines have been shown to predict the outcome of games more accurately many statistical models~\cite{song2007comparative}. This list of features will be modified \textit{a posteriori} to improve its performance for the next season~(in future work: Sec.~\ref{sec:conc}).

Relevant data were collected, organized, and utilized to make the FPTS predictions. Both player and team data were collected from the 2018 NFL regular season over 17 weeks. Data were collected from multiple sources shown in Table~\ref{tab:websites}. Data from these sources were aggregated into Excel and then arranged based on player and week of play. A master list was compiled from DraftKings draftable~(i.e.,~have an assigned salary for the upcoming game) players. 

\begin{table}[!!!htb]
\centering
\caption{Player and Team Data Resources. 
\vspace{0.1in}}\label{tab:websites}
\begin{tabular*}{13.0cm}{l|l}
\hline
Website & Data Procured \\ \hline
\href{hhtp://www.draftkings.com}{draftkings.com} & FPTS and Salary \\
\href{http://rotoguru1.com/cgi-bin/fyday.pl?week=1&game=dk}{rotoguru1.com} & Player Name, Position, Team, Team Schedule \\
\href{https://fantasydata.com/nfl-stats/point-spreads-and-odds?season=2017&seasontype=1&week=1}{fantasydata.com} & Vegas Lines, Over/Under, and Spread \\
\href{https://fusiontables.google.com/DataSource?dsrcid=110830#rows:id=1}{fusiontables.google.com} & Latitude and Longitude of Games \\
\href{https://www.footballoutsiders.com/dvoa-ratings/2017/week-1-dvoa-ratings
}{footballoutsiders.com} & Team, Offensive, and Defensive Ranking \\
\href{http://www.aussportsbetting.com/data/historical-nfl-results-and-odds-data/}{aussportsbetting.com} & Points Scored \\
\href{https://www.foxsports.com/nfl/gallery/2017-nfl-schedules-bye-week-dallas-cowboys-green-bay-packers-new-england-patriots-042117}{foxsports.com} & Team Bye Week \\ \hline
\end{tabular*}
\end{table}

Player positions were assigned a binary categorical value. Assigning integer values~(e.g.,~1-5) to positions, may mislead the NN as there would be the implication these values are \textit{ordinal}. Instead, positions were assigned five columns of categorical binary values for a QB, RB, WR, TE, and DST. Each player was given a respective ``1'' in the corresponding position column or a ``0'' otherwise. 

Four-week moving windows were created to use recent data for training the NN. These windows refer to a fraction of the season's data: For example, ``window one'' includes data from weeks one through four and ``window two'' includes data from weeks two through five. In total, fourteen windows span a season. To implement this moving window there must be at least four weeks of games played. These four weeks of play allow for the training of a NN to use with the upcoming window~(Sec.~\ref{sec:exp}). For the purpose of training the NN, players must have earned FPTS for game four of a window. This will ensure the NN has a target value. For players to be included in a window, they must have played at least four games over the past six weeks and must be draftable. These six weeks allow players to have two weeks of data replaced due to bye weeks or minor injuries. Increasing the window size will ultimately cut the player pool for lineup generation as it increases the proportion of players who have missed more than two games. To train a NN, two sets of data were made for each window including the features and target data. The features include the known data before the start of game four in a window, and the target data are the actual FPTS scored in the last game of a window. 

\begin{table}[!!!ht]
\centering
\caption{Training Data: Neural network training data set for each player including 43 features and 1 target. Data sets are categorized by two input data sets of a target that is represented by features. Each variable can be identified as being of a certain data type real, integer~(hierarchical), or binary~(categorical). Variable Number represent the number and location of variables in each data set. Game 4 refers to the \textit{upcoming} week of matches while Games 1-3 refer to the three previous weeks. \vspace{0.1in}}\label{tab:training}
\begin{tabular}{|l|l|l|l|}
\hline
\multicolumn{2}{|c|}{Data Set} & {Variable Number} & {Data Type}\\
\hline
\multirow{12}{*}{Features} & Position~(5 categories) & 1-5 & Binary\\
 & Game 1-3 FPTS & 6-8 & Real\\
 & Game 1-3 Point Differentials & 9-11 & Integer\\
 & Game 1-3 Team Offensive Rank & 12-14 & Integer\\ 
 & Game 1-3 Team Defensive Rank & 15-17 & Integer\\
 & Game 1-3 Opponent Offensive Rank & 18-20 & Integer\\
 & Game 1-3 Opponent Defensive Rank & 21-23 & Integer\\
 & Game 1-4 Home or Away & 24-27 & Binary\\ 
 & Game 1-4 Point Spread & 28-31 & Real\\
 & Game 1-4 Over/Under & 32-35 & Real\\
 & Game 1-4 Latitude & 36-39 & Real\\
 & Game 1-4 Longitude & 40-43 & Real\\ \hline
Target & Game 4 FPTS & 1 & Real\\ \hline
\end{tabular}
\end{table}

\subsection{Calculating expected fantasy points}\label{sec:exp} 

A NN was trained to predict the FPTS for each player in the upcoming Thursday through Monday games using the data collected and organized in Sec.~\ref{sec:agg}. A NN is a type of machine learning process loosely based on the behavior of neurons in the brain. Very simply, machine learning is a regression technique capable of myriad relations between the input and the output variables. NNs can be composed of different architectures that vary the structure and the method in which data are analyzed. A two-layer feed-forward neural network is made up of three layers: an input layer, one hidden layer, and an output layer. Layers in a NN are made up of ``neurons''~(also referred to as ``nodes'' or ``units'') that connect each neuron to the subsequent layer's neurons. Both the hidden and output layer perform a weighted average based on the weights and biases assigned to each neuron. The hidden layer uses an activation function that accounts for the non-linearity of the input data and feeds it into the output layer. Similarly, the output layer uses a linear function to yield an output. It ``learns'' by progressively adjusting connections between neurons to learn from the given data. Here, a supervised learning approach guides the network to a specific solution by defining ``features'' (input) and a ``target'' (output) data set. 

Features and their corresponding target as defined by the user are separated into three sets a training, validation, and testing set. To decrease MSE between the target and outputs, the NN adjusts the weights and biases according to the learning algorithm used and obtains a new output for both the training and validation set. It progressively adjusts the weights until the network stops improving on the validation set. The model may then be used with new data to determine an output when given the same set of features. 



\begin{figure}[t!!]
 \centering
 \includegraphics[width=0.99\linewidth]{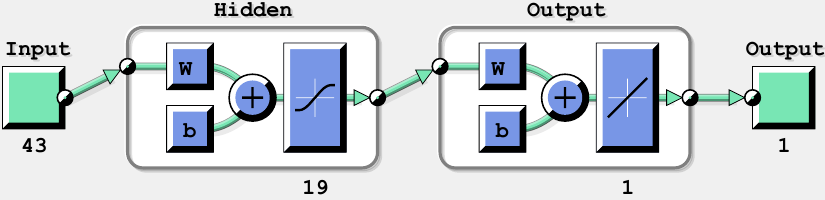}
 \caption{Schematic of Neural Network architecture. This model used 43 features with a single hidden layer of 19 neurons to output 1 FPTS value for the upcoming week. 
 }
 \label{fig:net}
\end{figure}

For this particular model, the features are the 43 listed and described in Table~\ref{tab:training}, and the one output is the FPTS for the upcoming game. Some features may be redundant individually but may help the model when correlations are made to other features~\cite{guyon2003introduction}. The training window includes all known data before the start of a game~(features) and the outcome player performance of the game~(target). NN modeling was implemented using \texttt{nftool} in Matlab. The toolbox implements a two-layer feed-forward network with one 19-node hidden layer and an output layer. The hidden layer uses sigmoid hidden neurons, and the output layer uses a linear output. To determine performance, the tool uses mean squared error~(MSE) between the true target and output from the network-in-training. A network schematic is shown in Fig.~\ref{fig:net}.

The final parameters and, consequently, the accuracy of the trained NN model are sensitive to a number of factors. These include, but are not limited to: the number of hidden layers, number of neurons, selected training algorithm, the ratio of training- to validation data, and which data are included in the training set versus the validation set. Some mixtures of the data may yield better predictions, on average, of the output than others. When making relationships among input data for a specific response, the most prominent nonlinear NN technique uses hidden layers trained with back-propagation of error~\cite{abdi1994neural}. One hidden layer of 19 neurons and the Bayesian regularization backpropagation learning algorithm~\cite{mackay1992bayesian} was employed for this NN model. In small-scale tests, too few or too many neurons degraded performance by under- or over-fitting the data. Another learning algorithm, Levenberg-Marquardt backpropagation~\cite{marquardt1963algorithm}, ran faster but did not perform as well. The model was created using 80\% of the data for training and 20\% for validation. To predict FPTS for week five, the NN is trained with window one features and target from weeks one through four. The testing set uses weeks two through five features only from window two. These features allow the trained NN to predict the FPTS of game four~(week five) in window two.


\subsection{Lineup generation}\label{sec:milp}

With the NN model now trained, FPTS for each draftable player in their upcoming game can be predicted. In this step, the lineup that maximizes the expected FPTS and adheres to the salary and position constraints is generated. For this contest~--~as introduced in Sec.~\ref{sec:intro}~--~the total salary may be at most \$50,000 with a nine-player lineup consisting of one quarterback~(QB), two running backs~(RB), three wide receivers~(WR), one tight end~(TE), one flex~(may be a RB, WR, or TE), and one defensive special team~(DST). A mixed integer linear programming~(MILP) is used for this maximization task. 

The MILP finds the values for a vector of binaries, $\textbf{x}$, that maximize the expected FPTS of the lineup. A value of ``1'' at the $i^{th}$ position in $\textbf{x}$ designates the $i^{th}$ player on the list being selected for the lineup while ``0'' means not selected. Vectors $\textbf{f}$ and $\textbf{S}$, corresponding to the predicted FPTS and salary of each player, respectively, were created. The dot product of these vectors and $\textbf{x}$ are calculated to find the lineup's predicted FPTS and salary. Position constraints were enforced by ensuring that the total selected players for each position add up to the required number. The objective function and constraints are shown mathematically in Eq.~\eqref{eq:milp}.

\begin{equation}\label{eq:milp}
 \underset{\mathbf{x}}{min} \; (-\mathbf{f}^T \mathbf{x}) \hspace{0.1in} \text{subject to} \; 
 \left\{
 \begin{array}{l}
 x_i\in \{0,1\} \\ \vspace{0.06in}
 \mathbf{S}^T \mathbf{x} \leq \$50,000 \\ \vspace{0.06in}
 \sum\limits_{i \in QB} x_i = 1 \\ \vspace{0.06in}
 \sum\limits_{i \in RB} x_i = n_{RB} \\ \vspace{0.06in}
 \sum\limits_{i \in WR} x_i = n_{WR} \\ \vspace{0.06in}
 \sum\limits_{i \in TE} x_i = n_{TE} \\ \vspace{0.06in}
 \sum\limits_{i \in DST} x_i = 1
 \end{array}
 \right.
 \end{equation}
 
The MILP was run three times: once for each possibility of the flex player's position~(RB, WR, or TE) by changing the values of $n$ each iteration as follows:

$$n_{RB} = 2, \; n_{WR} = 3, \; n_{TE} = 2, $$
$$n_{RB} = 2, \; n_{WR} = 4, \; n_{TE} = 1, $$
$$n_{RB} = 3, \; n_{WR} = 3, \; n_{TE} = 1. $$

The optimization was implemented using the \texttt{intlinprog} function in Matlab. The lineup that produced the maximum predicted FPTS of the three was selected as the optimal lineup. Three \texttt{intlinprog} options were used with the function: \texttt{AbsoluteGapTolerance} of $0$, \texttt{CutMaxIterations} of $25$, and \texttt{IntegerTolerance} of $1\times10^{-4}$.

Note that DraftKings requires players in a lineup be selected from at least two teams. This requirement was \textit{not} enforced in this step so the generation of infeasible lineups would be possible. 

\subsection{Statistical methods}\label{sec:stats}

To guard against the possibility of using a model with poor predictive ability due to a specific mix of the training and validation data, 10,000 models were created by placing data randomly into the training set or validation set for each instance in Sec.~\ref{sec:exp}. Data for the upcoming week was then placed into each model, creating 10,000 FPTS predictions for each draftable player. Using the FPTS from one model at a time, 10,000 lineups were generated using the MILP procedure in Sec.~\ref{sec:milp}. 

As only one lineup will be entered into the contest, the most robust lineup of the 10,000 was selected as the final choice. Here, the most robust lineup was chosen to be the modal lineup: the one that was generated most often among the simulations. Using the modal roster is analogous to using a majority/plurality voting~\cite{hansen1990neural} scheme in classification problems. In convergence testing, by 100 lineups, the modal lineup converged and was unchanged when adding more simulations. 

Confidence intervals~(CI) of the percentile of performance~(Secs.~\ref{sec:random} and \ref{sec:real}) were determined using bootstrapping analysis~\cite{efron1992bootstrap} with a resample size of 10,000. 

Normality of distributions was tested using a one-sample Kolmogorov-Smirnov test. Means of distributions were compared using a two-sided unpaired, heteroscedastic t-test. The significance threshold was set \textit{a priori} to $\alpha=0.05$. Effect size between the means of distributions was calculated using Cohen's $d$~\cite{cohen1988statistical}. Because of the normality of the distributions and large sample size, no corrections were performed. 

\section{Results}\label{sec:results}

\subsection{Weekly generated lineups}\label{sec:weekly}

For each week of games, data from the previous three weeks were used in the trained NN model. Data from the active week~(not the FPTS) were also included. These data were processed by the methods described in Secs.~\ref{sec:exp} and \ref{sec:milp}. Players declared on injured reserve, ``out'', ``questionable'' or otherwise not starting were manually removed from the possible selections. Player salaries and assigned positions were taken from DraftKings 

\begin{figure}[t!!]
 \centering
 \includegraphics[width=0.99\linewidth]{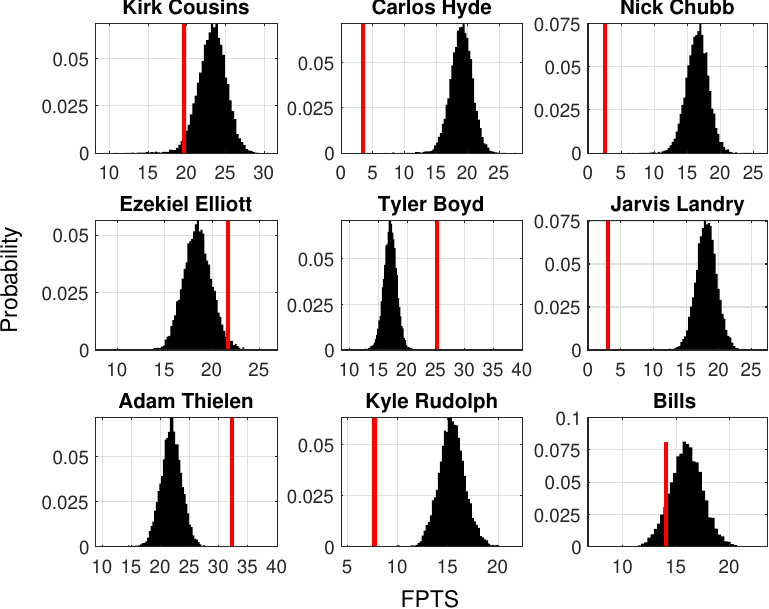}
 \caption[Line Up 6]{Probability histograms of expected player FPTS for generated lineup for week 6. The FPTS Distributions are based on 10,000 models. Total FPTS \emph{predicted} to be 165.6~[145.8,~185.5]. The \emph{actual} total FPTS was 129.52. Red lines indicate a player's actual FPTS.} 
 \label{fig:lineup6}
\end{figure}

Lineups were generated in real-time for NFL weeks 6 through 16 before the start of each week's first game. In weeks 7 and 9, a player in the generated lineup did not play in the game~(as a game-time decision after the lineup was set) and thus generated zero FPTS. Therefore, these weeks were not included in the analysis in the following sections. The predicted and actual performances of generated lineups are summarized in Table~\ref{tab:lineups}. An example lineup from week 6 is shown in Fig.~\ref{fig:lineup6}. All generated lineups are shown in Sec.~\ref{sec:genlineups}.

\begin{table}[!!!ht]
\centering
\caption[Lineup Performance]{Predicted and Actual FPTS Performance of Weekly Generated Lineups. The drafted players are seen in the associated histograms. Note, weeks 7 and 9 each had a game-time substitution where a drafted player did not participate. These weeks were not included for consideration. \vspace{0.1in}}\label{tab:lineups}
\begin{tabular}{|c|c|r|l|}
\hline
\multicolumn{1}{|c}{Week} & \multicolumn{1}{|c}{Predicted FPTS [95\% CI]} & \multicolumn{1}{|c|}{Actual FPTS} & \multicolumn{1}{c|}{Lineup Histogram} \\ \hline
6 & 165.6~[145.8,~185.5] & 129.52 & Fig.~\ref{fig:lineup6} \\ \hline
8 & 167.0~[145.9,~185.0] & 134.70 & Fig.~\ref{fig:lineup8} \\ \hline
10 & 160.1~[137.6,~186.6] & 120.80 & Fig.~\ref{fig:lineup10} \\ \hline
11 & 174.1~[153.1,~195.7] & 125.10 & Fig.~\ref{fig:lineup11} \\ \hline
12 & 185.7~[157.8,~213.5] & 136.96 & Fig.~\ref{fig:lineup12} \\ \hline
13 & 187.0~[169.1,~205.2] & 95.88 & Fig.~\ref{fig:lineup13} \\ \hline
14 & 145.7~[127.8,~165.5] & 88.80 & Fig.~\ref{fig:lineup14} \\ \hline
15 & 193.8~[161.8,~222.9] & 100.12 & Fig.~\ref{fig:lineup15} \\ \hline
16 & 135.7~[118.5,~154.5] & 94.54 & Fig.~\ref{fig:lineup16} \\ \hline
\end{tabular}
\end{table}


\subsection{Comparison to randomly-created lineups}\label{sec:random}

Each week, the generated lineups from Sec.~\ref{sec:weekly} were first tested against 35,000~(approximately the same number that will be used in Sec.~\ref{sec:real}) feasible, \textit{randomly}-created lineups. These lineups used a total salary of at least \$45,000~--~90\% of the cap~--~to serve as a reasonable proxy for competing with low-skill users. This is similar to the method used by Newell~\cite{newell2017optimizing} that compares optimal lineups generated by an integer program to randomly generated lineups using at least 90\% of the salary cap. The random lineups were created after the conclusion of the final game each week and did not use players that earned zero FPTS to avoid drafting any player that did not play or earn FPTS that week. 

\begin{figure}[t!!]
 \centering
 \includegraphics[width=0.999\linewidth]{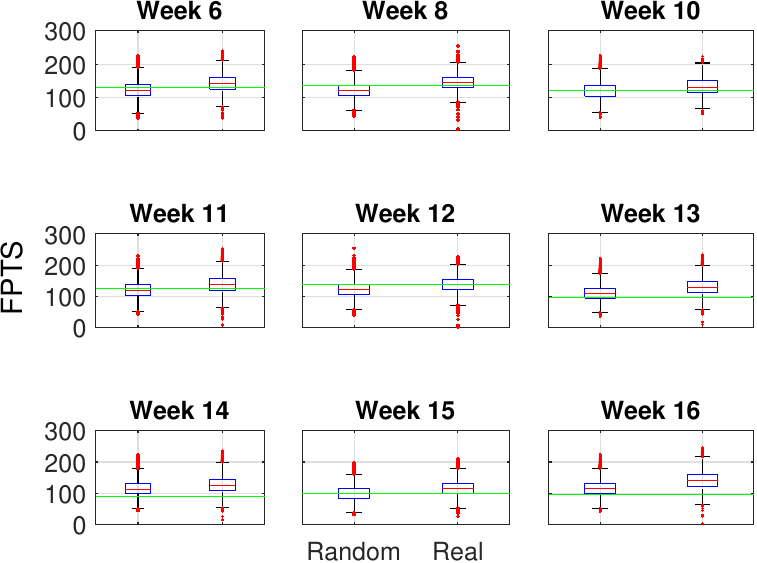}
 \caption[Boxplots]{Boxplots of weekly FPTS distributions of randomly-created lineups (left box in each) and real-word (right box in each). The mean  of the real-world distributions was found to be larger than the mean of the randomly-created with all $p \ll 0.001$. The green line indicates the number of FPTS scored by the generated lineup in each week.} 
 \label{fig:rand2real}
\end{figure}

The performance of each week's lineup was evaluated by calculating its FPTS percentile within the distribution of random lineups. Boxplots of the distribution of random lineups compared to the generated lineup are shown per week in Fig.~\ref{fig:rand2real} and the calculated percentiles are shown in Table~\ref{tab:random}. The randomly-created lineup FPTS distributions are the left-sided boxes in each plot. The FPTS distributions of every week passed the Kolmogorov-Smirnov normality test with $p \ll 0.001$.

\begin{table}[!!!ht!!!!]
\centering
\caption[Performance versus Random]{Weekly Generated Lineup Performance Against Randomly-Created Lineups. Percentile of the generated lineup within the random lineups and 95\% CI is shown. 
\vspace{0.1in}}\label{tab:random}
\begin{tabular}{|c|r|c|}
\hline
\multicolumn{1}{|c}{Week} & \multicolumn{1}{|c|}{FPTS} & \multicolumn{1}{c|}{Percentile [95\% CI]} \\ \hline
6 & 129.52 & 62.7~[62.2,~63.2] \\ \hline
8 & 134.70 & 72.6~[72.1,~73.1] \\ \hline
10 & 120.80 & 51.4~[50.9,~51.9] \\ \hline
11 & 125.10 & 57.8~[57.3,~58.3] \\ \hline
12 & 136.96 & 73.4~[73.0,~73.9] \\ \hline
13 & 95.88 & 27.3~[26.8,~27.7] \\ \hline
14 & 88.80 & 13.6~[13.3,~14.0] \\ \hline
15 & 100.12 & 52.9~[52.3,~53.4] \\ \hline
16 & 94.54 & 19.3~[18.9,~19.7] \\ \hline
\end{tabular}
\end{table}

\subsection{Comparison to real-world DFS contests}\label{sec:real}

To next test the performance of the method, the generated lineups were submitted and compared to those from real DFS users. The lineups and FPTS totals for all users were collected from GPP contests on DraftKings using the NFL ``Classic'' Thursday through Monday format. All contests were either free or \$0.25 to enter. This contest uses every game for a week of regular season play. The results from several contests each week were collected and aggregated. Week six of the regular season was used as the starting point for entry and data collection as the methodology requires four weeks of past data for the training the NN model. Also, by this week, the starters for the teams are likely determined. User scores of zero were removed from the data set when calculating the percentiles and displaying the histograms. The summary of weekly performances is shown in Table~\ref{tab:retro}. Boxplots of the distributions of the FPTS of the real-world users and the FPTS of the generated lineup are shown in Fig.~\ref{fig:rand2real}. The real-world FPTS distributions are the right-sided boxes in each plot. The FPTS distributions of every week, except week 6, passed the Kolmogorov-Smirnov normality test with $p < 0.05$. 

Every week, the mean FPTS of the real-world distribution was larger than the mean of the random distribution with $p \ll 0.001$. The effect sizes ranged from $d=0.49$ to $d=1.08$, indicating medium to large effect sizes every week that was tested. 

\begin{table}[!!!ht] 
\centering
\caption[Generated Lineup versus Real-World]{Weekly Lineup Performance Against Real-World Contests. Percentile of generated lineup in the user lineups and 95\% CI is shown. Note, week 10 had a relatively low sample size. 
\vspace{0.1in}}\label{tab:retro}
\begin{tabular}{|c|r|r|r|r|}
\hline
\multicolumn{1}{|c}{Week} & \multicolumn{1}{|c}{FPTS} & \multicolumn{1}{|c}{FPTS of Users [95\% CI]} & \multicolumn{1}{|c|}{Percentile [95\% CI]} & \multicolumn{1}{c|}{Users} \\ \hline
6 & 129.52 & 141.0~[93.21,~190.9] & 32.2~[31.5,~32.9] & 16,500 \\ \hline
8 & 134.70 & 145.2~[104.3,~189.3] & 31.6~[31.1,~32.0] & 37,300 \\ \hline
10 & 120.80 & 130.3~[87.97,~189.0] & 34.2~[32.3,~36.2] & 2300 \\ \hline
11 & 125.10 & 136.8~[84.68,~195.0] & 33.1~[32.6,~33.7] & 29,000 \\ \hline
12 & 136.96 & 136.7~[92.34,~185.0] & 50.5~[49.8,~51.2] & 22,000 \\ \hline
13 & 95.88 & 128.8~[81.78,~180.7] & 9.34~[8.99,~9.70] & 25,000 \\ \hline
14 & 88.80 & 125.4~[80.15,~183.0] & 6.10~[5.82,~6.40] & 26,200 \\ \hline
15 & 100.12 & 115.3~[72.02,~164.3] & 25.2~[24.7,~25.6] & 33,000 \\ \hline
16 & 94.54 & 139.6~[88.22,~195.9] & 4.52~[4.24,~4.81] & 20,400 \\ \hline
\end{tabular}
\end{table}

\section{Discussion}\label{sec:disc}

The generated lineups outperformed randomly-created lineups, on average, using the simulation process outlined in Sec.~\ref{sec:random}. The generated lineup received FPTS typically above the median with the exception of weeks 13, 14, and 16. However, the generated lineups did not fare as well against real-world users. As shown in Sec.~\ref{sec:real}, the generated lineups had a median performance at the 31\textsuperscript{st}-percentile. The methodology consistently outperformed about one-third of human participants.

The actual FPTS earned by the generated lineups each week fell below the low-end 2.5$^{th}$-percentile prediction from the models. At the \textit{player} level, individual FPTS predictions were generally poor, usually falling far outside the 95\% CI from the 10,000 models. This can be seen visually in Appendix~\ref{sec:genlineups}. As mentioned in Sec.~\ref{sec:exp}, the predictive ability of the NN is sensitive to the training algorithm, the number of neurons, the training and validation set size and ratio, and the selected features. There are many parameters within the NN that may be adjusted and features may be varied to increase model performance. The accuracy of the model is greatly dependent on the feature variables, so determining the variables that have the most influence on the target is crucial. Some other features that were tested include a player's team and opponents. These were used as categorical features but dramatically increased the computational time by adding 160 features. The MILP was incentivized to choose the ``undervalued'' players according to the NN prediction. However, the unusually high FPTS prediction for some players may have been due to modeling error rather than true insight discovered from the NN. 

The performance of the generated lineups took a noticeable decline starting in week 13 against both the random lineups and real-world contests. There are multiple possible contributions for this time-based decline in performance. First, Cumulative injury may also contribute to the reduced prediction power as the season evolves. Though players continue to start, injury prevalence increases in the second half of the season~\cite{binney}. Consequently, players may not be playing as well as earlier weeks of the season. Second, around this week in the season, teams begin to clinch playoff spots or be eliminated from post-season contention. High-performing players may be held back in these instances to mitigate their risk of injury before the post-season. Betting against NFL teams that have clinched playoffs, has been shown in the literature to be a successful strategy~\cite{krieger2015anchoring}. Conversely, as teams are eliminated from the post-season, there is an incentive to continue to lose to obtain a higher draft pick in the upcoming draft~\cite{tuck2013lead}. The concept is known as ``tanking'' and is expected to strengthen a team in the upcoming season. Neither of these possible degradation factors is considered by the NN. As the current model looks back three weeks, it is not well suited to adjust for these abrupt changes in player and team performance. One remedy for these potential issues would be to enter contests only in the middle weeks of the season.



User skill has been shown to be necessary for DFS success~\cite{eastondaily}. After submitting random lineups to 35 MLB double-up contests and losing on all instances~\cite{newell2017optimizing}. In their study, random lineups represent unskilled users and utilize only feasible lineups without a minimum salary. On average, these lineups ranked in the bottom 6.12\% of contest lineups. This supports that simply selecting players blindly for a lineup is not a sufficient strategy for success in DFS contests. Based on the t-test results in Sec.~\ref{sec:real}, the mean FPTS from user lineups was significantly higher than the mean of the randomly-created lineups. All effect sizes were at least ``medium'' between the two groups. These results are consistent with previous work, demonstrating that skillfully selecting DFS lineups outperforms random selections.
Assuming a normal distribution of MLB contest results, a comparison to NFL DFS contests, on average optimized lineups ranked in the bottom 25.2\%: nearly a 312\% increase in percentile from random lineups. By simply having a strategy for selecting a lineup, the percentile of the lineup may increase drastically. 

The same study also simulated weekly FanDuel NFL DFS contests by generating feasible random lineups with a 90\% minimum salary to represent real contest users as a comparison against optimized lineups representing a skilled user. In this case, the skilled participant is maximizing expected FPTS by using an integer program. The average FPTS of past games were used with the simulation. Their results found~--~with 99\% confidence~--~that the random team's mean FPTS will be 20 to 55 points lower than that of an integer program. The results presented in Sec.~\ref{sec:random} offer additional support that lineups can be generated that outperform random selections. For most weeks, the lineups were above the median. The data collected from DraftKings in Sec.~\ref{sec:real} show too that users perform better than random rosters. 

The methodology was validated against only one format of many possible Daily Fantasy games. On DraftKings, it can be used directly for the smaller pools~(e.g.,~Sunday- or Monday-only contests) by using the same model but limiting the players eligible for selection in the MILP~(Sec.~\ref{sec:milp}) step. Different sites~(e.g.,~FanDuel, FantasyDraft) can be used by simply changing the source of the FPTS data. With modifications, other sports could be used with the methodology. 


The premise of this method's MILP optimization was to maximize expected \textit{FPTS}. Alternatively, the optimization could be changed to maximize expected \textit{winnings}. As the GPP awards to a top performing percentile of lineups, one does not need to be the best to win a payout. In two DraftKings NFL DFS contest studies~\cite{newell2017optimizing2, newell2017optimizing}, a stochastic integer program is expected to have a higher payout in ``tiered'' contests with lineups having a lower mean and higher standard deviation. Conversely, the integer program produced lineups with a higher mean and lower standard deviation and is expected to have a higher payout in a ``double-up'' contest. Tiered and double-up refer to different contests where tiered contests payout about 20\% of users and double up contests payout 50\%, but both followed the classic-style constraints considered in this work. These findings coincide with salary allocation strategies for different payout structures, suggesting a more volatile hit-or-miss lineup has a higher probability of receiving a payout in GPP structures~\cite{HowtoAll38:online}. Similarly, a more consistent lineup is expected to do better in double-up structures. As shown in Sec.~\ref{sec:real}, the selected players generally underperformed the prediction, but also show volatility which may be beneficial for GPP contests.

Although GPP contests range in allowable entries per room from one to unlimited, only one lineup was submitted to one contest every week. A study of a DraftKings high-stakes contest showed that nearly all participants that submitted fewer than 100 entries lost money~\cite{Natta_2016}. Many of the top users submit \textit{multiple} lineups. However, such a strategy would not be satisfactory for contests with only one permissible entry. This methodology currently produces one lineup, but it could be used to identify multiple potential lineups which could be entered. 



Real-world validation was performed using the data from the combination of multiple free or \$0.25 contests. Data from pay-to-play contests were not available from DraftKings unless a lineup was entered. It is conceivable that the caliber of the user and the selection strategy of a lineup is different in free and low-cost contests versus higher-fee contests which may skew the FPTS distributions. 

Due to computational and data constraints, model training was limited. Running the training for longer and more epochs may improve the results. Using more features could also improve its predictive power. However, including categorical data, like teams, may add 32 more binary features for each instance. The number of features versus the amount of data may cause an overfit system and the training time increases dramatically. 
The training window data for the NN was set to three previous weeks in this study. Using more or fewer weeks in the window may improve the accuracy of the FPTS prediction. Future work will permute and compare many different modeling options to improve the outcomes.

\section{Conclusions \& future work}\label{sec:conc}



The FPTS of players is difficult to predict, but statistical modeling of historical data allows for forecasting of player FPTS. Predicting player FPTS is further shadowed by the uncertainty involved with chance. Injuries and days players simply underperform are difficult or nearly impossible to predict even with all available quantified player data. DFS is growing in popularity, and as more users join the online contests some may be skeptical as to the gambling aspect of fantasy sports. However, with statistical modeling, there is mathematical logic behind selecting lineups and avoids human bias. Statistical models attempt to determine undervalued players that most users may overlook. This method may be applied to other SKP with dynamic resource values with a desire to allocate resources optimally. The player estimations using the model have a high margin of error, but it is believed the error may be reduced substantially with the right combination of features and functions within the NN. 

Unlike some previous FPTS prediction studies, this method does not assume a normal distribution of FPTS nor the independence of players' FPTS. The use of the NN modeling allows for the discovery of ``synergy'' between players on the same team. 

The authors did not have experience with fantasy football prior to the study and present an objective methodology and analysis. With this baseline study complete, experienced fantasy users will be brought in to refine the process. Future improvements in performance can be compared to this work. 

First, post hoc factor analysis will be performed on the NN models to reveal what features were most and least salient to the FPTS prediction. The low-contributing factors may be pruned: leading to a speedup in the training time without an appreciable loss in accuracy. Furthermore, reducing the feature set will help avoid overfitting the data. There is a relatively small amount of data, ranging from 150-300 players in the window. Replacing low-contributing features with new higher-contributing features may increase the fidelity while minimizing overfitting.

Weights and biases on features going to neurons can be given initial values, so adjusting weights may put emphasis on certain variables to start training. Network functions may be varied to select training and validation set data differently, or adjusting the pre-processing of features and post-processing of outputs may also reduce error. If the model does not produce reliable results in the future, a different machine learning algorithm may be applied. The NN model generally overestimates player FPTS, but it may be adjusted for further reduction in estimation error. 

This methodology could be extended to other NFL DFS styles, such as ``Showdown.'' Here, users draft a six-player team of all flex players with one being the captain with a 50\% increase in salary but a 50\% bonus in FPTS. 

Other sports could benefit from the principles introduced by this method. For example, MLB lineups could be generated and could have better performance due to an increased data set size with over ten times as many games in the regular season per team compared to the NFL.


\section*{Acknowledgments}

The authors would like to thank the Penn State Multi-Campus Research Experience for Undergraduates for financial support. Thanks to Dr.~Chris Byrne for his insights and suggestions to the methodology and analysis. 

\bibliography{draft.bib}
\bibliographystyle{unsrt} 

\renewcommand{\thesection}{\Alph{section}}
\setcounter{section}{0}

\section{Selected players and FPTS prediction histograms of generated weekly lineups}\label{sec:genlineups}


\begin{figure}[p!!]
 \centering
 \includegraphics[width=0.99\linewidth]{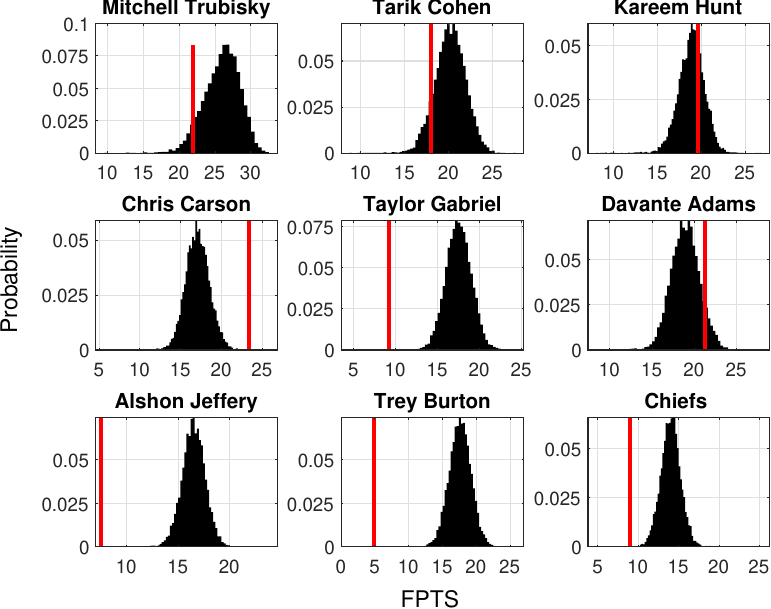}
 \caption[Line Up 8]{Probability histograms of expected FPTS for generated lineup for week 8. Distributions based on 10,000 models. Total FPTS \emph{predicted} to be 167.0~[145.9,~185.0]. The \emph{actual} FPTS was 134.7. Red lines indicate the player's actual FPTS.} 
 \label{fig:lineup8}
\end{figure}


\begin{figure}[p!!]
 \centering
 \includegraphics[width=0.99\linewidth]{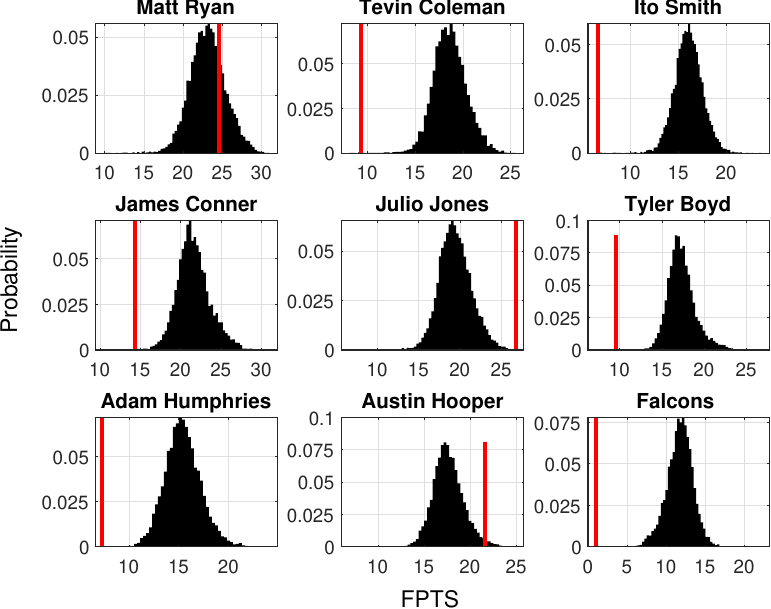}
 \caption[Line Up 10]{Probability histograms of expected FPTS for generated lineup for week 10. Distributions based on 10,000 models. Total FPTS \emph{predicted} to be 160.1~[137.6,~186.6]. The \emph{actual} FPTS was 120.8. Red lines indicate the player's actual FPTS.}
 \label{fig:lineup10}
\end{figure}


\begin{figure}[p!!]
 \centering
 \includegraphics[width=0.99\linewidth]{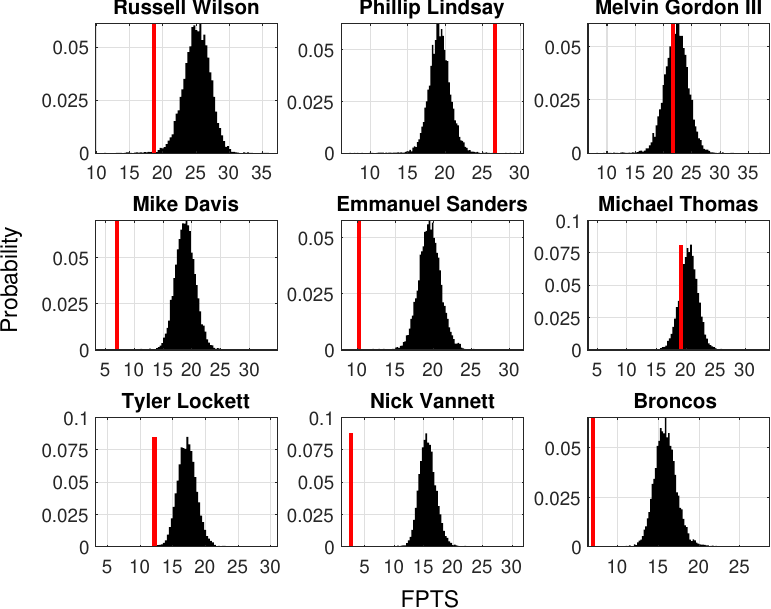}
 \caption[Line Up 11]{Probability histograms of expected FPTS for generated lineup for week 11. Distributions based on 10,000 models. Total FPTS \emph{predicted} to be 174.1~[153.1,~195.7]. The \emph{actual} FPTS was 125.1. Red lines indicate the player's actual FPTS.}
 \label{fig:lineup11}
\end{figure}


\begin{figure}[p!!]
 \centering
 \includegraphics[width=0.99\linewidth]{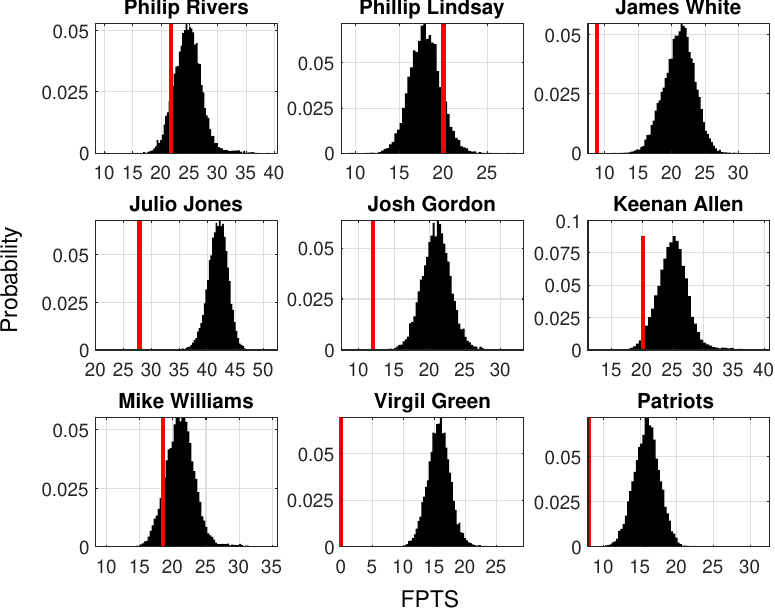}
 \caption[Line Up 12]{Probability histograms of expected FPTS for generated lineup for week 12. Distributions based on 10,000 models. Total FPTS \emph{predicted} to be 184.6~[158.0,~210.9]. The \emph{actual} FPTS was 136.96. Red lines indicate the player's actual FPTS.}
 \label{fig:lineup12}
\end{figure}


\begin{figure}[p!!]
 \centering
 \includegraphics[width=0.99\linewidth]{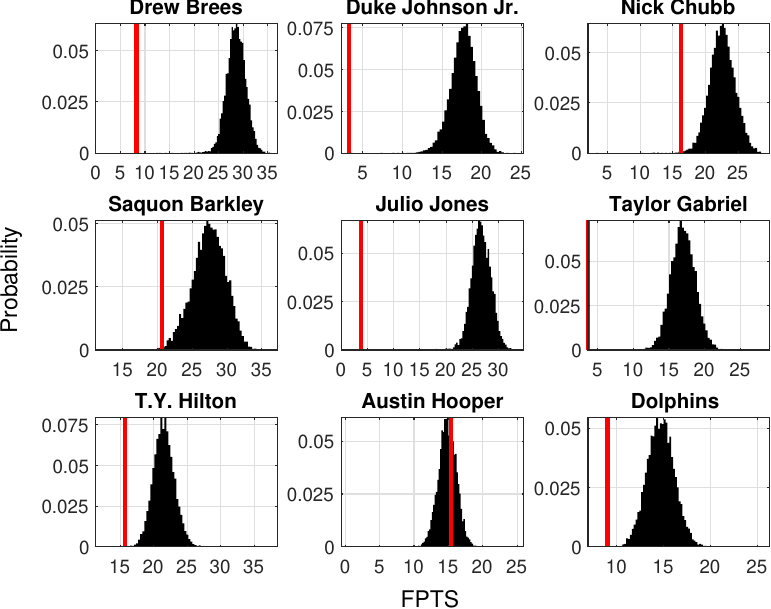}
 \caption[Line Up 13]{Probability histograms of expected FPTS for generated lineup for week 13. Distributions based on 10,000 models. Total FPTS \emph{predicted} to be 187.0~[169.1,~205.2]. The \emph{actual} FPTS was 95.88. Red lines indicate the player's actual FPTS.}
 \label{fig:lineup13}
\end{figure}


\begin{figure}[p!!]
 \centering
 \includegraphics[width=0.99\linewidth]{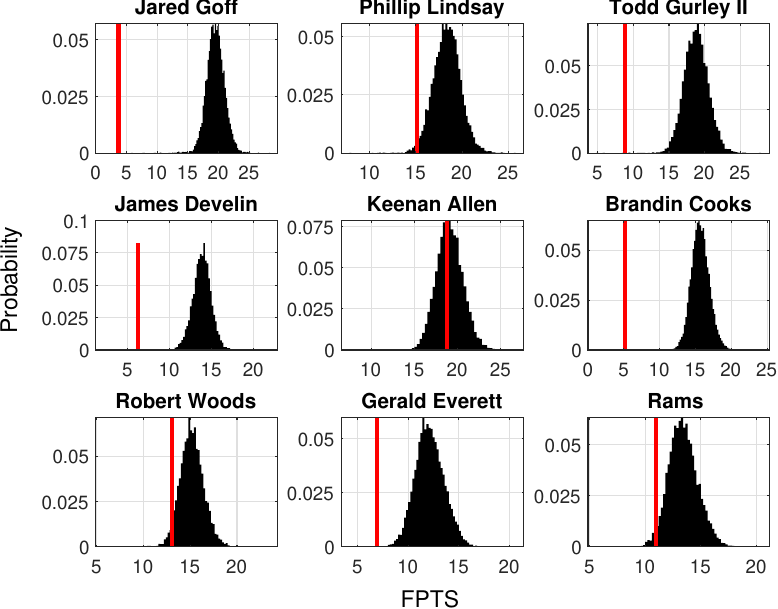}
 \caption[Line Up 14]{Probability histograms of expected FPTS for generated lineup for week 14. Distributions based on 10,000 models. Total FPTS \emph{predicted} to be 145.7~[127.8,~165.5]. The \emph{actual} FPTS was 88.8. Red lines indicate the player's actual FPTS.}
 \label{fig:lineup14}
\end{figure}


\begin{figure}[p!!]
 \centering
 \includegraphics[width=0.99\linewidth]{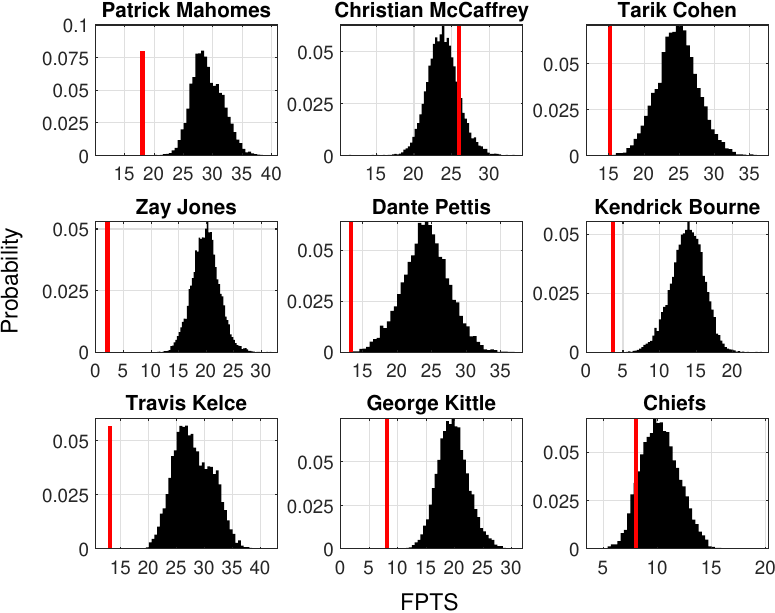}
 \caption[Line Up 15]{Probability histograms of expected FPTS for generated lineup for week 15. Distributions based on 10,000 models. Total FPTS \emph{predicted} to be 193.8~[161.8,~222.9]. The \emph{actual} FPTS was 100.12. Red lines indicate the player's actual FPTS.}
 \label{fig:lineup15}
\end{figure}


\begin{figure}[p!!]
 \centering
 \includegraphics[width=0.99\linewidth]{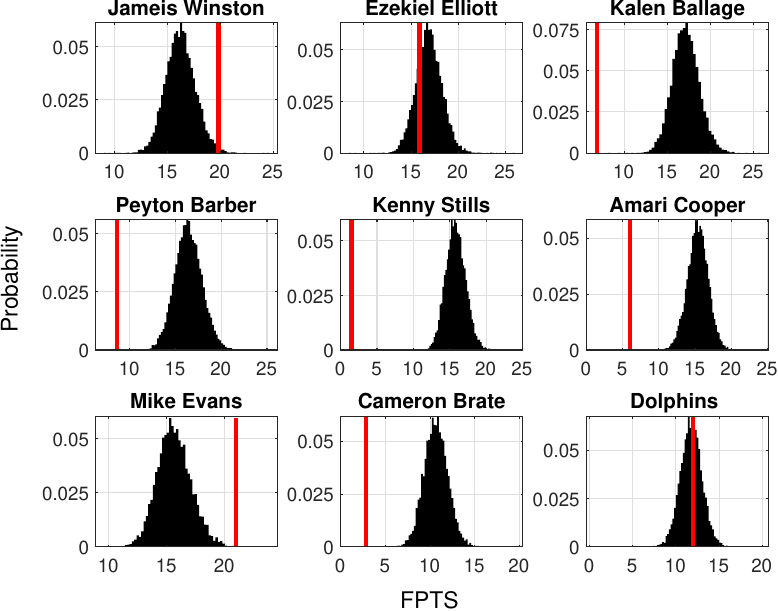}
 \caption[Line Up 16]{Probability histograms of expected FPTS for generated lineup for week 16. Distributions based on 10,000 models. Total FPTS \emph{predicted} to be 135.7~[118.5,~154.5]. The \emph{actual} FPTS was 94.54. Red lines indicate the player's actual FPTS.}
 \label{fig:lineup16}
\end{figure}

\end{document}